\documentclass[10pt,journal]{IEEEtran}
\usepackage{times}
\usepackage{epsfig}
\usepackage{graphicx}
\usepackage{amssymb}
\usepackage{multirow}
\usepackage{enumitem}
\usepackage{xspace}
\usepackage{cite}
\usepackage[cmex10]{amsmath}
\usepackage{caption}
\usepackage{url}

\makeatletter
\DeclareRobustCommand\onedot{\futurelet\@let@token\@onedot}
\def\@onedot{\ifx\@let@token.\else.\null\fi\xspace}

\def\eg{\emph{e.g}\onedot} 
\def\ie{\emph{i.e}\onedot} 
 
\def\etc{\emph{etc}\onedot} 
 
\def\etal{\emph{et al}\onedot}
\makeatother

\begin{document}

\title{Skeleton Based Action Recognition Using Translation-Scale Invariant Image Mapping And Multi-Scale Deep CNN}

\author{
Bo Li$^{1}$,
Mingyi He$^1$
Xuelian Cheng$^1$
Yucheng Chen$^1$
Yuchao Dai$^{1}$,

$^1$ Northwestern Polytechnical University, China
}

\maketitle

\begin{abstract}
This paper presents an image classification based approach for skeleton-based video action recognition problem. Firstly, A dataset independent translation-scale invariant image mapping method is proposed, which transformes the skeleton videos to colour images, named skeleton-images. Secondly, A multi-scale deep convolutional neural network (CNN) architecture is proposed which could be built  and fine-tuned on the powerful pre-trained CNNs, \eg, AlexNet, VGGNet, ResNet \etal. Even though the skeleton-images are very different from natural images, the fine-tune strategy still works well. 
At last, we prove that our method could also work well on 2D skeleton video data. We achieve the state-of-the-art results on the popular benchmard  datasets \eg NTU RGB+D, UTD-MHAD, MSRC-12, and G3D. Especially on the largest and challenge NTU RGB+D, UTD-MHAD, and MSRC-12 dataset, our method outperforms other methods by a large margion, which proves the efficacy of the proposed method.

\end{abstract}

\section{Introduction}
\label{sec:intro}
Action recognition is an important research area in computer vision, which has a wide range of applications, \eg, human computer interaction, video surveillance, robotics, and \etc. Recent years, the cost-effective depth sensor combining with real-time skeleton estimation algorithms can provide reliable joint coordinates \cite{Shotton2013Efficient}. As an intrinsic high level representation, 3D skeleton is valuable and comprehensive for summarizing a series of human dynamics in the video, and thus benefits general action analysis \cite{Liu2017PKU}. Besides its succinctness and effectiveness, it has a significant advantage of great robustness to illumination, clustered background, and camera motion. Based on these advantages, 3D skeleton based activity analysis has drawn great attentions \cite{Zanfir2013The, Gowayyed2013Histogram, Vemulapalli2014Human, Vemulapalli2016Rolling, Ohnbar2013Joint, Du2015Hierarchical, Veeriah2015Differential, Zhu2016Co, Shahroudy2016NTU, Liu2016Spatio, Song2016An,Du2016Representation}.

Very recently, human pose estimation from 2D RGB videos have also been studied with deep CNN method~\cite{Zhou2015Sparseness,Yang2016End,Toshev2013DeepPose}. Rather accurate 2D skeleton joints could be evaluated from RGB videos. However, it seems still lack effective method to deal with this kind of 2D skeleton video data for recognition purpose.

Previously, hand-crafted skeleton features have been devised~\cite{ Wu2014Leveraging, Vemulapalli2014Human, Barnachon2014Ongoing, Vemulapalli2016R3DG, Vemulapalli2016Rolling}. However, these hand-crafted features are always shallow and dataset-dependent, thus limiting their performance.

Recently, deep learning based methods have achieved great success in high-level computer vision tasks such as image recognition, classification, detection, and semantic segmentation \etc. As for the 3D skeleton based action recognition problem, Recurrent Neural Networks (RNNs) have been widely adopted \cite{Du2015Hierarchical, Veeriah2015Differential, Zhu2016Co, Shahroudy2016NTU, Liu2016Spatio, Song2016An, Du2016Representation}. RNNs could effectively extract the temporal information and learn the contextural information well. However, RNNs tend to overemphasize the temporal information especially when the training data is insufficient, thus leading to over-fitting~\cite{Wang2016Action}.

Convolutional Neural Networks (CNNs) have also been applied to this problem \cite{Wang2016Action, Yong2015Skeleton}. Different from the RNNs, how to effectively represent 3D skeleton data and feed into deep CNNs is still an open problem. 
Wang \etal \cite{Wang2016Action} proposed the Joint Trajectory Maps (JTM), which represents both spatial configuration and dynamics of joint trajectories into three texture images through color encoding, and then fed these texture images to CNNs for classification. However, this kind of JTM is a little complicated, and may lose some important information when projecting 3D skeleton into 2D image. Du \etal \cite{Yong2015Skeleton} proposed to represent each skeleton sequence as an image, where the temporal dynamics of the sequence are encoded as changes in columns and the spatial structure of each frame is represented as column. Their encoding method is dataset dependent and translation-scale variant which means their encoding method need dataset information and human's translation and action scale may influence the final mapping results.

\begin{figure*}[!htb]
\centering
\includegraphics[width=0.85\linewidth]{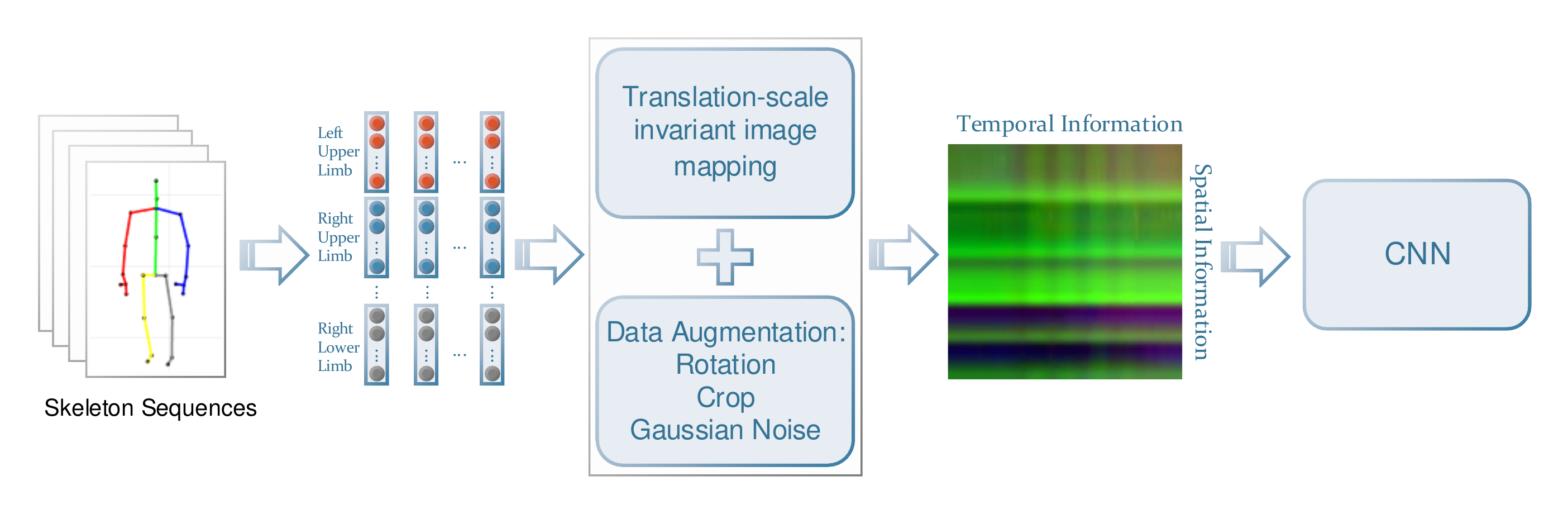}
\caption{Flowchart of the proposed method. The 3D skeleton video is first mapped to action image via translation-scale invariant transformation. Image classification is conducted on the action image by using our multi-scale CNNs.}
\label{fig:flowchart}
\end{figure*}

To tackle the above shortcomings in skeleton-based video recognition problem, in this paper, we present a new framework consisting of translation-scale invariant image mapping and multi-scale deep CNN classifier. The overall flowchart of our method is illustrated in Fig.~\ref{fig:flowchart}. We propose to map the 3D skeleton video to a color image, where the color image achieves translation and scale invariance and dataset independent. The proposed mapping could easily handle the translation and scale changes in 3D skeleton data, thus is more distinctive, and dataset independent. 

Although the skeleton images are very different from natural images, the widely used pre-trained deep CNN model \eg, AlexNet\cite{krizhevsky2012imagenet}, VGGNet\cite{Simonyan2014Very}, ResNet\cite{ResNet} could still be transferred to it well. It is especially valuable when there is insufficient annotated skeleton videos. The fine-tune strategy could avoid training millions of parameters afresh and improve the performance significantly~\cite{Wang2016Action, Girshick2013Rich}. More importantly, due to the special property of this kind of skeleton images, we propose a simple yet effective multi-scale deep CNN to enhance the frequency adjustment ability of our method. 

In addition, we extend our method to deal with 2D skeleton-based video recognition problem. Surprisingly, our method could also work well. Experimental results on the popular benchmark dataset like NTU RGB-D, UTD-MHAD, MSRC-12, and G3D demonstrate the effectiveness our proposed framework. We also give extensive analysis experiments to show the propoerties of our method.

In conclusion, our main contributions are summarized as following:
\begin{itemize}
  \item We propose a translation-scale invariant image mapping method for 3D skeleton video data. This mapping could avoid the translation and scale influence of the skeleton data, thus is more distinctive and dataset independent.
  

  \item A multi-scale deep CNN is proposed to enhance the frequency adjustment ability of our method.
  
  \item We test our method to 2D skeleton data and achieve excellent results which shows our method could also work on 2D skeleton data well.

  \item We achieve the state-of-the-art results on the widely used benchmarks like NTU RGB-D, UTD-MHAD, MSRC-12, and G3D dataset. In addition, extensive component analysis experiments are conducted.
\end{itemize}

This paper are organized as following. Related works are summarized and presented in Sec.~\ref{rel-work}. We present our method in Sec.~\ref{method}, including our Translation-scale invariant image mapping method, multi-scale deep CNN and data augmentation method. Experiments on the popular benchmarks are presented in Sec.~\ref{experiments}. At last, more analysis experiments and results on 2D skeleton data are presented in Sec.~\ref{analysis}.


\section{Related Work}\label{rel-work}

Tranditionaly, hand-crafted skeleton features have been devised to capture the spatial-tempory information. There have been great amount of hand-craft features proposed for 3D skeleton-based video recognition~\cite{ Wu2014Leveraging, Vemulapalli2014Human, Barnachon2014Ongoing, Vemulapalli2016R3DG, Vemulapalli2016Rolling}. Generally speaking, spatial descriptor, geometric descriptor, or key poses are extensively studied. We would like suggest the readers refer to \cite{Presti20163D} for more summary. However, these hand-crafted features are always shallow and dataset-dependent, thus limiting their performance.

Recently, deep learning methods have been adopted on this field. Du \etal~\cite{Du2015Hierarchical} divided the human skeleton into 5 parts and a hierarchical recurrent neural network is proposed for this problem. Veeriah \etal~\cite{Veeriah2015Differential} proposed a kind of differential recurrent neural network which emphasizes on the change in information gain caused by the salient motions between the successive frames. Zhu \etal~\cite{Zhu2016Co} proposed a kind of regularized deep LSTM network which take the skeleton as the input at each time slot and introduce a novel regularization scheme to learn the co-occurrence features of skeleton joints. Liu \etal~\cite{Liu2016Spatio} propose a more powerful tree-structure based traversal method. To handle the noise and occlusion in 3D skeleton data, they introduce new gating mechanism within LSTM to learn the reliability of the sequential input data. 

Convolutional Neural Networks (CNNs) have also been applied to this problem \cite{Wang2016Action, Yong2015Skeleton}. Wang \etal \cite{Wang2016Action} proposed the Joint Trajectory Maps (JTM), which represents both spatial configuration and dynamics of joint trajectories into three texture images through color encoding, and then fed these texture images to CNNs for classification. Du \etal \cite{Yong2015Skeleton} proposed to represent each skeleton sequence as an image, where the temporal dynamics of the sequence are encoded as changes in columns and the spatial structure of each frame is represented as column. How to effectively represent 3D skeleton data and feed into deep CNNs is still an open problem.

In recent years, human pose estimation from 2D RGB videos have been studied with deep CNN method~\cite{Zhou2015Sparseness,Yang2016End,Toshev2013DeepPose}. Rather accurate 2D skeleton joints could be evaluated from RGB videos.

Our work is also related to recent works on transfer learning and deep learning. In~\cite{krizhevsky2012imagenet}, Krizhevsky \etal trained a large deep CNN on the ImageNet dataset and achieved a performance leap on the classification problem. Recently, more and more work show that pre-trained deep CNN features can be transferred to new classification or recognition problems and boost performance~\cite{Girshick2013Rich, oquab2013learning}. In~\cite{Simonyan2014Very}, a very deep CNN is proposed, which achieved state-of-the-art classification results in ImageNet challenge 2014. In~\cite{ResNet}, a kind of residual structure is proposed which makes the hundreds layers CNN trainable. 

This paper is an extended version of our conference paper published in ICMEW~\cite{}, In which, we achieve the 3rd place in the ``Large Scale 3D Human Activity Analysis Challenge in Depth Videos".  However, this journal version substantively improves the performance and gives more insights by more experiments and analysis.

\section{Method}\label{method}
Our framework consist of 3 parts (1) Translation-scale invariant image mapping. (2) A multi-scale deep CNN for classification. (3) Data augmentation methods we utilized for 3D skeleton data. A conceptual illustration of our framework is presented in Fig.\ref{fig:flowchart}.

\subsection{Translation-scale invariant image mapping}
Following the work of \cite{Yong2015Skeleton}, we divide all human skeleton joints in each frame into five main parts according to human physical structure, \ie two arms, two legs and a trunk. To preserve the local motion characteristics, joints in each part are concatenated as a vector by their physical connections. Then the five parts are concatenated as the representation of each frame. To map the 3D skeleton video to an image, a natural and direct way is to represent the three coordinate components $(x,y,z)$ of each joint as the corresponding three components $(R,G,B)$ of each pixel in an image. Specially, each row of the action image is defined as $R_i = [x_{i1},x_{i2},...,x_{iN}]$, $G_i = [y_{i1},y_{i2},...,y_{iN}]$, $B_i = [z_{i1},z_{i2},...,z_{iN}]$, where $i$ denotes the joint index and $N$ indicates the number of frames in a sequence. By concatenating all joints together, we obtain the resultant action image representation of the original skeleton video.

Due to the coordinate difference in 3D skeleton and image, proper normalization is needed. Du \etal \cite{Yong2015Skeleton} proposed to quantify the float matrix to discrete image representation with respect to the entire training dataset. Specifically, given the joint coordinate $c^{jk}$ ($x,y,z$), the corresponding pixel value $p^{jk}$ is defined as
\begin{equation}
\label{softmax}
p^{jk} = floor\left(255 * \frac{c^{jk}-c_{min}}{c_{max}-c_{min}}\right),
\end{equation}
where $c_{max}$ and $c_{min}$ are the maximum and minimum of all joint coordinates in the training set respectively. $jk$ represent the $k$-th channel ($x,y,z$) of the $j$-th skeleton video sequence. $floor$ is the rounding down function. However, as the normalization is conducted with respect to the entire training dataset, the resultant action image is dataset dependent and translation-scale variant.  1. This encoding method need $c_{min}, c_{max}$ which need statistic on the dataset. Different dataset may have different $c_{min}, c_{max}$, thus this kind of encoding method is dataset dependent. 2. The same action conducted on the different position(caused by the translation of the person or camera) will result in a different action image, thus is translation variant. 3. The same action conducted by different subjects may have some scale variant, while this normalization also could not guarantee scale invariant in encoding the skeleton video.

\begin{figure}[htb]
\centering
\includegraphics[width=0.85\linewidth]{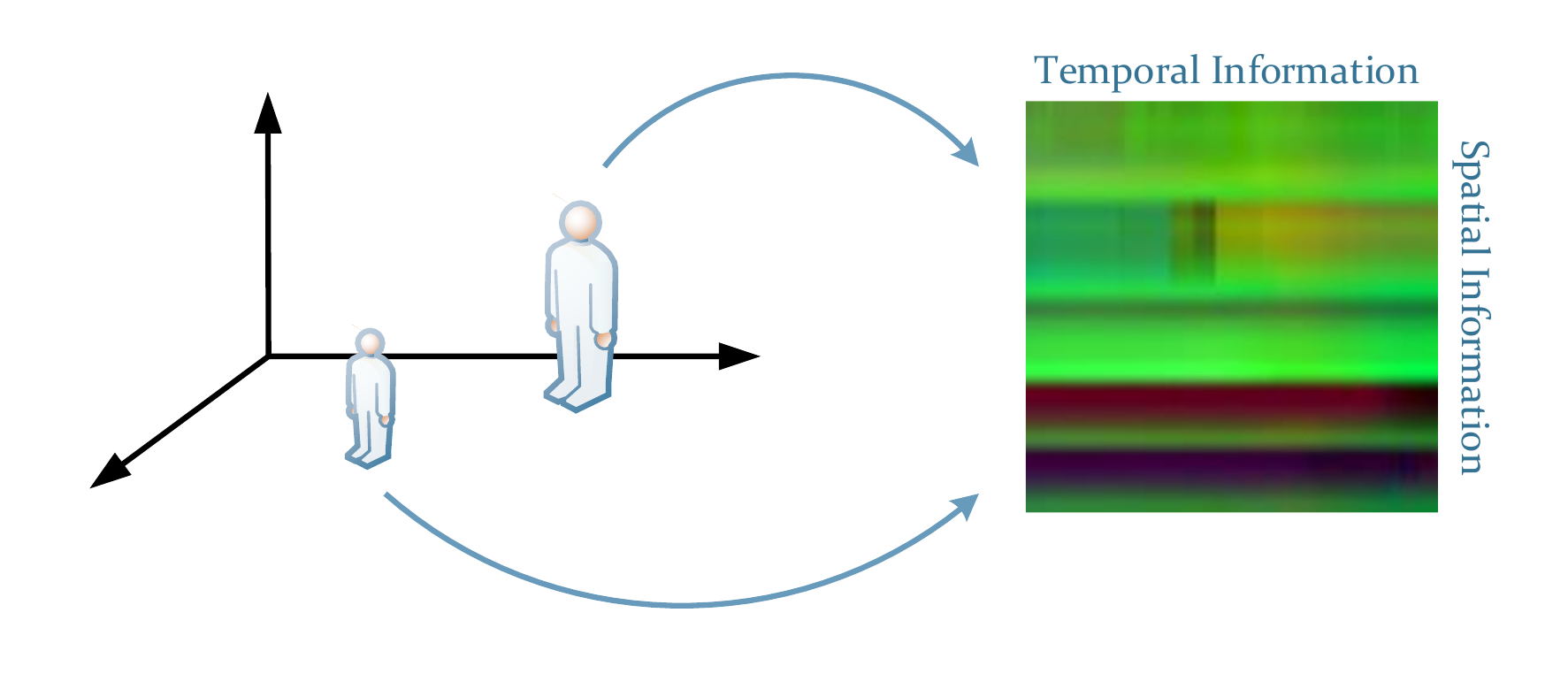}
\caption{Illustration of the translation-scale invariant image mapping.}
\label{fig:map}
\end{figure}

To tackle the above problems in encoding the skeleton video, we propose a simple and yet effective translation-scale invariant image mapping method, which is presented in Equ.~\ref{eq:our_mapping}
\begin{equation}
\label{eq:our_mapping}
p^{jk} = floor\left(255 * \frac{c^{jk}-c^{jk}_{min}}{\max\limits_{k}(c^{jk}_{max}-c^{jk}_{min})}\right),
\end{equation}
where $c^{jk}_{max}$ and $c^{jk}_{min}$ are the maximum and minimum coordinate value of the $k$-th channel ($x,y,z$) of the $j$-th skeleton video sequence.

Compared with Du \etal \cite{Yong2015Skeleton}, our image mapping method owns the following properties:
\begin{itemize}
  \item \textbf{Translation invariant:} Our image mapping transforms the 3D coordinates with respect to the minimum coordinate in each sequence rather than the entire training dataset, thus the translation in 3D does not affect the action video as illustrated in Fig.~\ref{fig:map}.
  \item \textbf{Scale invariant:} Similarly, by normalizing the 3D skeleton coordinates with respect to the 3D variation along the three axis in each sequence, the scale change has also been eliminated. In addition, our normalization is isometric to each coordinate $x,y,z$, thus the relative scale in different axis has been well preserved.
  \item \textbf{Dataset independent:} As the minimum and maximum coordinates are extracted from each sequence independently, the normalization is thus independent to each specific skeleton video dataset. 
\end{itemize}

\begin{figure}[htb]
\centering
\scalebox{0.9}{
\begin{tabular}{@{}c@{}c@{}c}
\includegraphics[width=0.45\linewidth]{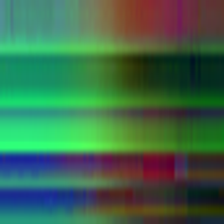} \ \
&\includegraphics[width=0.45\linewidth]{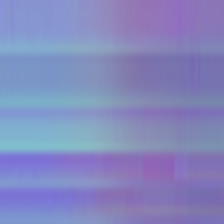}\\
Our & Du \etal～\cite{Yong2015Skeleton} \\
\end{tabular}
}
\caption{Comparison between the mapping results from \cite{Yong2015Skeleton} and our method.}
\end{figure}

\subsection{Action recognition through multi-scale CNN}

\begin{figure}[htb]
\centering
\includegraphics[width=0.85\linewidth]{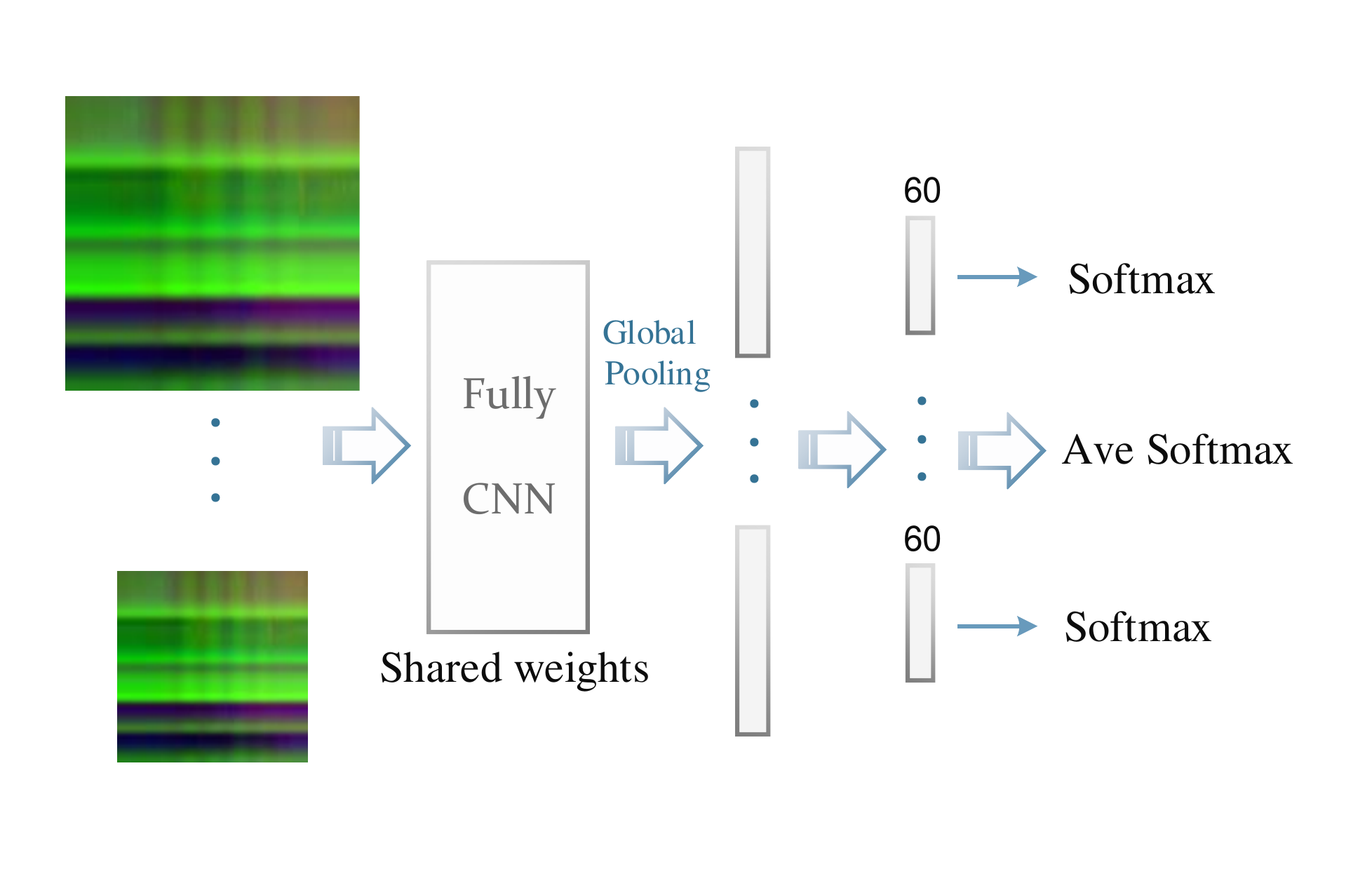}
\caption{Our multi-scale pretrained CNN model.}
\label{fig:multi-scale}
\end{figure}

Our overall multi-scale CNN architecture is presented in Fig.~\ref{fig:multi-scale}. Our CNN architecture could be built on the fully-CNN based pre-trained model like AlexNet\cite{krizhevsky2012imagenet}, VGGNet\cite{Simonyan2014Very}, and ResNet~\cite{ResNet}.

The multi-scale structure is motivated by the frequency variant of this kind of skeleton-images. Under our skeleton mapping framework, if we fix the size of the convolution kernel, different input size will bring different frequency variance. The multi-scale (multi-frequency) input often includes rich cues to the activity recognition problem.

In order to reduce the amount of the parameters in our model, the weights of the Fully-CNN parts are shared by all the different resolution inputs. Then global pooling is performed on the correspondent feature maps which is critical to result in the same size feature vectors. In order to further regularize the training, we put on the softmax loss on the all output of different resolution as well as the average of them.

We train our network with the multinomial logistic loss:

\begin{equation}
\begin{aligned} 
E = -\frac{1}{N} \sum_{n=1}^N {\log(p_{k}^n)}
\end{aligned}
\end{equation}

\noindent $p$ is the output of the softmax layer:

\begin{equation}
\begin{aligned}
p_{i} = \frac {\exp{x_{i}}} {\sum_{i^{'}=1}^{m}{\exp{x_{i^{'}}}}}
\end{aligned}
\end{equation}

\noindent Here, $N$ is the number of training samples, $k$ the correspondent label of sample $n$, $m$ is the number of classes. $x$ is the input of the softmax layer.

\subsection{Data augmentation}
Data augmentation has been proved as an effective way in deep CNN based image classification. In this paper, we have encoded the 3D skeleton data to RGB images. In order to augment the dataset and leap the classification performance, we have specially designed different data augmentation strategies, such as 3D coordinate random rotation~\cite{Wang2016Action}, Gaussian noise, video crop etc.

\begin{figure}[htb]
\centering
\scalebox{0.9}{
\begin{tabular}{@{}c@{}c@{}c@{}c@{}c}
\includegraphics[width=0.19\linewidth]{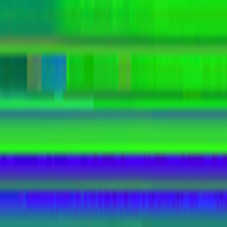} \ \
&\includegraphics[width=0.19\linewidth]{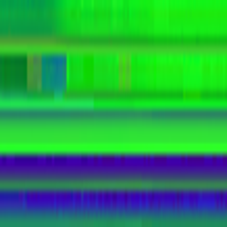} \ \
&\includegraphics[width=0.19\linewidth]{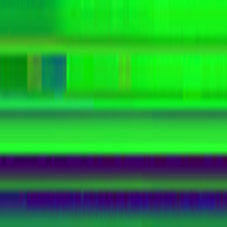} \ \
&\includegraphics[width=0.19\linewidth]{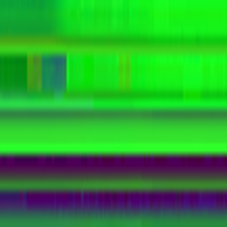} \ \
&\includegraphics[width=0.19\linewidth]{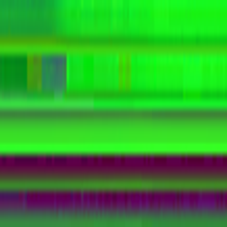} \\
(a) &(b) &(c) &(d) &(e)
\end{tabular}
}
\caption{Illustration of skeleton image results from different rotation settings along $x$ axis. (a) $-30^o$, (b) $-15^o$, (c) $0^o$, (d) $15^o$, (e) $30^o$ mapping results.}
\label{fig:rot}
\end{figure}

The augmentation methods we utilized including:

\begin{itemize}

  \item 3D coordinate rotation: 3D coordinate are randomly rotated in the range of $[-30^o, 30^o]$, along the $x, y, z$ axis. Some examples are presented in Fig.~\ref{fig:rot}.

  \item Gaussian noise: We randomly add gaussian noise in the 3D coordinate with the $\theta = 0, \sigma = 0.01$. 

  \item Video crop: We randomly crop the videos by the range of $[0.7, 1]$ in the random locations of the video .

\end{itemize}

\section{Implementation details}
Before proceeding to the experimental results, we give implementation details for our method. 
The proposed network is trained by using stochastic gradient decent momentum of 0.9, and weight decay of 0.0004. Weights are initialized by the pre-trained model from AlexNet\cite{krizhevsky2012imagenet}, VGGNet\cite{Simonyan2014Very}, and ResNet~\cite{ResNet}. The network is trained with fixed learning rate 0.001 in the first 8 epoches, then divided by 10 every 5 epoches.

Our implementation is based on the efficient CNN toolbox: caffe~\cite{Jia2014Caffe} with an NVIDIA Tesla Titian X GPU.

\section{Experiments}
\label{experiments}
In this section, we evaluate the proposed method on public benchmark datasets: the large NTU RGB+D Dataset, UTD-MHAD, MSRC-12 Kinect Gesture Dataset, and G3D. The final recognition results were compared with the state-of theart reporteds on the same datasets. 

\subsection{NTU RGB+D Dataset}	
To the best of our knowledge, NTU RGB-D dataset \cite{Shahroudy2016NTU} is the largest action recognition dataset. We adopt the same train-test  protocol as in~\cite{Wang2016Action}.

The dataset has more than 56 thousands sequences and 4 million frames, containing 60 actions performed by 40 subjects aging between 10 and 35. It consists of front view, two side views and one left, right 45 degree views. This dataset is very challenging due to the large intra-class and viewpoint variations.

For a fair comparison and evaluation, the same protocol as that in \cite{Shahroudy2016NTU} was used. It has both cross-subject and cross view evaluation. In the cross-subject evaluation, samples of subjects 1,2,4,5,8,9,13,14,15,16,17,18,19,25,27,28,31,34,35 and 38 were used as training samples and the remaining subjects were reserved for testing. In the cross-view evaluation, samples taken by camera 2 and 3 were used as training, while testing set includes the samples of camera 1. We further augment the training dataset by 2 times. 

\begin{figure}[htb]
\centering
\scalebox{0.9}{
\begin{tabular}{@{}c@{}c@{}c}
\includegraphics[width=0.45\linewidth]{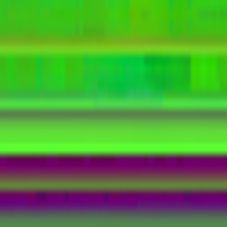} \ \
&\includegraphics[width=0.45\linewidth]{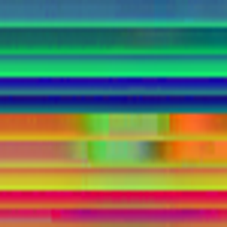}\\

1 person & 2 person \\
\end{tabular}
}
\caption{Illustration of the skeleton images for 1 person and 2 person in NTU RGB+D dataset}
\label{fig:NTUmap}
\end{figure}

In the dataset of NTU, there are some samples consisted of more than 1 person. We choose the simplest strategy to deal with this kind of situation, that, we just concat these two person's coordinate and present them in one image. An sample is presented in Fig.~\ref{fig:NTUmap}.

In Table \ref{tab:NTU}, we report the performance comparison between our method and the state-of-the-art methods. Clearly, our proposed method achieves the best performance in both cross-subject and cross-view evaluation. Our method outperforms the current state-of-the-art methods with a margin of 12\% in cross-subject evaluation and 11\% in cross-view evaluation.

\begin{table}[htb]
\begin{center}
\caption{Performance comparison on the NTU RGB-D dataset \cite{Shahroudy2016NTU}} \label{tab:NTU}
\begin{tabular}{  c  c  c }
\hline
Method              &Cross subject &Cross view \\
\hline
Lie Group~\cite{Vemulapalli2014Human} & 50.08\% & 52.76\% \\
Dynamic Skeletons~\cite{Ohnbar2013Joint} &60.23\% &65.22\% \\
HBRNN~\cite{Du2015Hierarchical} &59.07\% &63.97\% \\
Part-aware LSTM~\cite{Shahroudy2016NTU} &62.93\% &70.27\% \\
ST-LSTM + Trust Gate~\cite{Liu2016Spatio} &69.20\% & 77.70\% \\
JTM+CNN~\cite{Wang2016Action} & 76.32\% &81.08\% \\
STA-LSTM~\cite{Song2016An} &73.4\% &81.2\% \\
Proposed Method     & \bf{85.02\%} &\bf{92.3\%} \\
\hline
\end{tabular}
\end{center}
\end{table}

\subsection{UTD-MHAD}
UTD-MHAD \cite{Chen2015UTD} is a multimodal action dataset, captured
by one Microsoft Kinect camera and one wearable inertial
sensor. This dataset contains 27 actions performed by 8
subjects (4 females and 4 males) with each subject performing each action 4 times. After removing three corrupted sequences, the dataset has 861 sequences. The actions are: “right arm swipe to the left”, “right arm swipe to the right”, “right hand wave”, “two hand front clap”, “right arm throw”, “cross arms in the chest”, “basketball shoot”, “right hand draw x”, “right hand draw circle (clockwise)”, “right hand draw circle (counter clockwise)”, “draw triangle”, “bowling (right hand)”,
“front boxing”, “baseball swing from right”, “tennis right hand forehand swing”, “arm curl (two arms)”, “tennis serve”, “two hand push”, “right hand know on door”, “right hand catch an object”, “right hand pick up and throw”, “jogging in place”,“walking in place”, “sit to stand”, “stand to sit”, “forward lunge (left foot forward)” and “squat (two arms stretch out)”.It covers sport actions (e.g. “bowling”, “tennis serve” and “baseball swing”), hand gestures (e.g. “draw X”, “draw triangle”, and “draw circle”), daily activities (e.g. “knock on door”, “sit to stand” and “stand to sit”) and training exercises(e.g. “arm curl”, “lung” and “squat”). For this dataset, cross-subjects protocol was adopted as in \cite{Chen2015UTD}, namely, the data from the subjects numbered 1, 3, 5, 7 were used for training while subjects 2, 4, 6, 8 were used for testing. Table VIII compares the performance of the proposed method and those reported in \cite{Chen2015UTD}.

\begin{table}[htb]
\begin{center}
\caption{Performance comparison on the UTD-MHAD dataset \cite{Shahroudy2016NTU}} \label{tab:UTD-MHAD}
\begin{tabular}{  c  c }
\hline
Method              &Accuracy \\
\hline
ELC-KSVD~\cite{Zhou2014Discriminative} &76.19\%\\
kinect \& Inertial~\cite{Chen2015UTD} &79.10\%\\
Cov3DJ~\cite{Hussein2013Human}  &85.58\%\\
SOS~\cite{Hou2016Skeleton}  &86.97\%\\
JTM~\cite{Wang2016Action}& 87.90\%\\
Proposed Method     & \bf{96.27\%}\\
\hline
\end{tabular}
\end{center}
\end{table}

\begin{figure}[htb]
\centering
\includegraphics[width=0.95\linewidth]{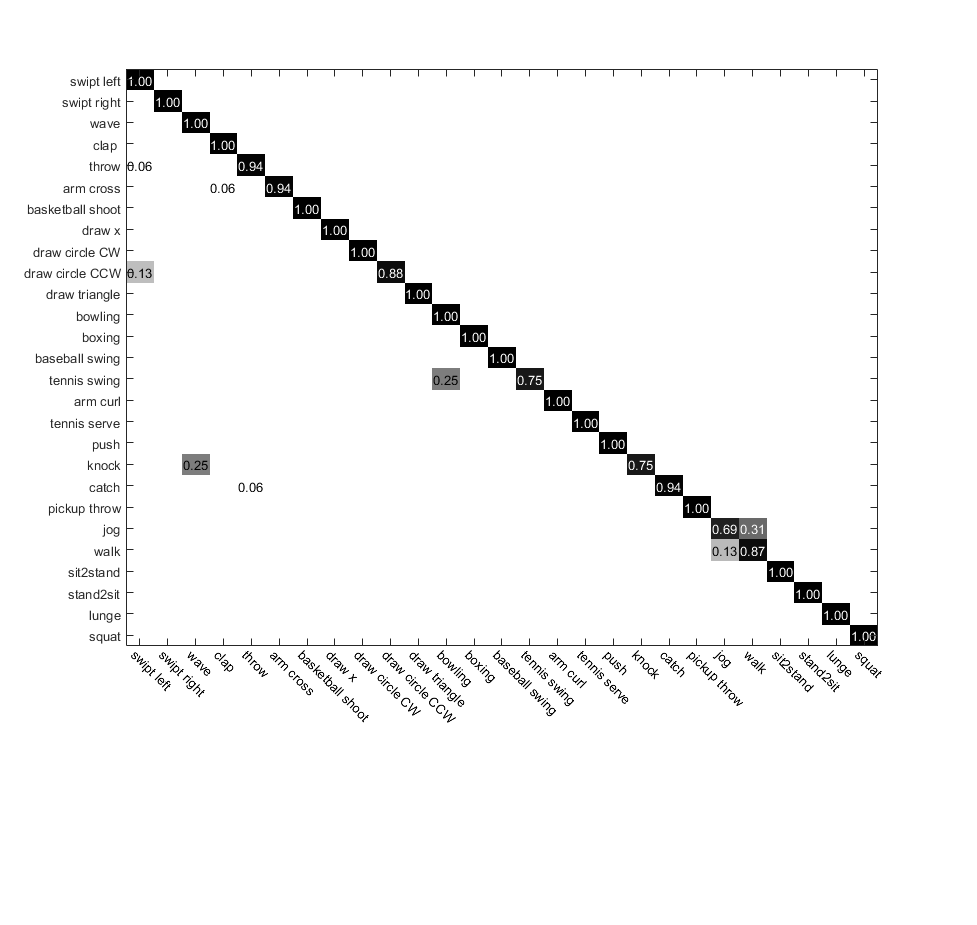}
\caption{Confusion matrix of UTD-MHAD}
\label{fig:map}
\end{figure}

\subsection{MSRC-12 Kinect Gesture Dataset}
MSRC-12 \cite{Fothergill2012Instructing} is a relatively large dataset for gesture/action
recognition from 3D skeleton data captured by a Kinect sensor. The dataset has 594 sequences, containing 12 gestures by 30 subjects, 6244 gesture instances in total. The 12 gestures are: “lift outstretched arms”, “duck”, “push right”, “goggles”,“wind it up”, “shoot”, “bow”, “throw”, “had enough”, “beat both”, ”change weapon” and “kick”. For this dataset, cross-subjects protocol was adopted, that is, odd subjects were used for training and even subjects were for testing.

\begin{table}[htb]
\begin{center}
\caption{Performance comparison on the MSRC-12 Kinect Gesture Dataset \cite{Shahroudy2016NTU}} \label{tab:MSRC-12}
\begin{tabular}{  c  c }
\hline
Method              &Accuracy \\
\hline
HGM~\cite{Yang2014A} &66.25\%\\
Pose-Lexicon~\cite{Zhou2016Learning} &85.86\%\\
ELC-KSVD~\cite{Zhou2014Discriminative} &90.22\%\\
Cov3DJ~\cite{Hussein2013Human}  &91.70\%\\
SOS~\cite{Hou2016Skeleton}  &94.27\%\\
JTM~\cite{Wang2016Action}& {94.86}\%\\
Proposed Method     & \bf{99.41\%}\\
\hline
\end{tabular}
\end{center}
\end{table}

\begin{figure}[htb]
\centering
\includegraphics[width=0.95\linewidth]{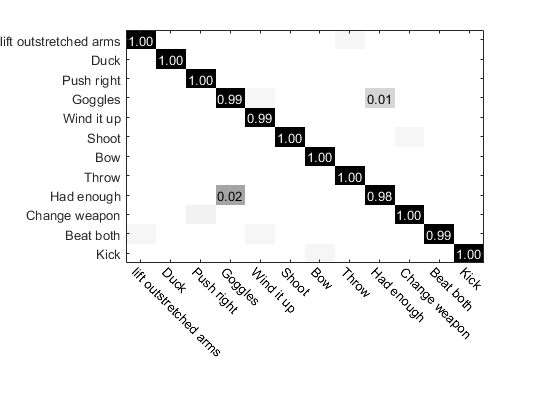}
\caption{Confusion matrix of MSRC 12 dataset}
\label{fig:map}
\end{figure}

\subsection{G3D Dataset}
Gaming 3D Dataset (G3D)\cite{Bloom2012G3D} focuses on real-time action recognition in a gaming scenario. It contains 10 subjects performing 20 gaming actions: “punch right”, “punch left”, “kick right”, “kick left”, “defend”, “golf swing”, “tennis swing forehand”, “tennis swing backhand”, “tennis serve”, “throw bowling ball”, “aim and fire gun”, “walk”, “run”, “jump”, “climb”, “crouch”, “steer a car”, “wave”, “flap” and “clap”. For this dataset, the first 4 subjects were used for training, the fifth for validation and the remaining 5 subjects were for testing as configured in \cite{Nie2015A}. Table VII compared the performance of the proposed method and those reported in \cite{Nie2015A}.

\begin{table}[htb]
\begin{center}
\caption{Performance comparison on the G3D dataset \cite{Shahroudy2016NTU}} \label{tab:G3D}
\begin{tabular}{  c  c }
\hline
Method              &Accuracy \\
\hline
ELC-KSVD~\cite{Zhou2014Discriminative} &82.37\%\\
Cov3DJ~\cite{Hussein2013Human}  &71.95\%\\
LRBM~\cite{Nie2015A} & 90.50\%\\
SOS~\cite{Hou2016Skeleton}  &95.45\%\\
JTM~\cite{Wang2016Action}& \bf{96.02}\%\\
Proposed Method     & {93.9\%}\\
\hline
\end{tabular}
\end{center}
\end{table}

\begin{figure}[htb]
\centering
\includegraphics[width=0.95\linewidth]{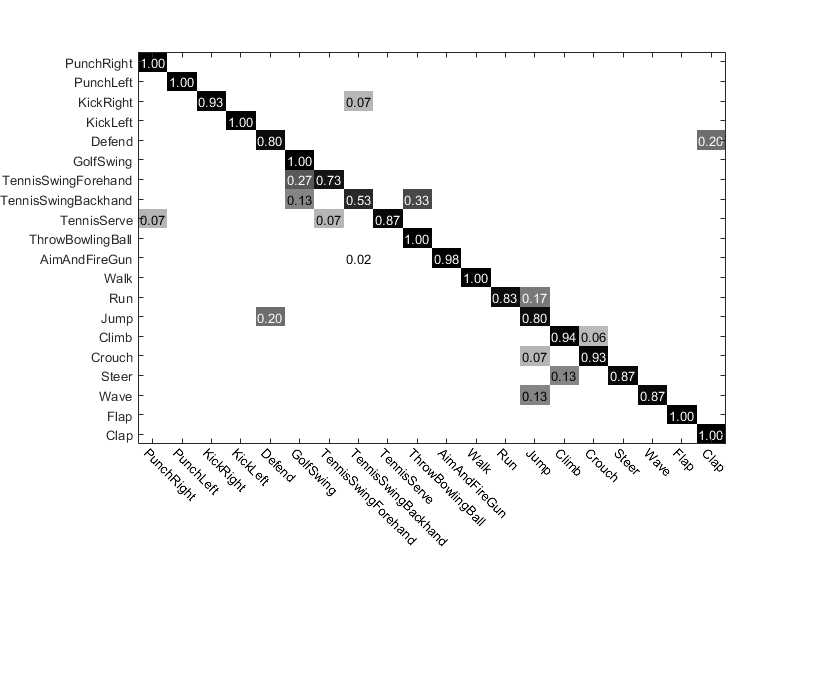}
\caption{Confusion matrix of G3D dataset}
\label{fig:map}
\end{figure}


\section{More analysis}
\label{analysis}

\subsection{Effect of our mapping method}

%
%
%
\begin{table*}[!htb]
\begin{center}
\caption{Performance comparison of different mapping methods on the popular benchmard dataset } \label{tab:analysis}
\begin{tabular}{  c  c  c  c  c  c}
\hline
Method           &NTU-CS &NTU-CV &UTD-MHAD &MSRC &G3D \\
\hline
Du~\cite{Yong2015Skeleton} &76.0\% &84.7\% &83.55\% &99.30\% &{92.5\%}\\
Our     & \bf{80.2}\% &\bf{85.0}\% &\bf{87.7}\% &\bf{99.4}\% &\bf{92.6}\%\\
\hline
JTM~\cite{Wang2016Action} + Aug  &76.3\% &81.1\% &87.90\% &{94.86}\% &\bf{96.02}\%\\
Our + Aug   & \bf{81.3}\% &\bf{87.2}\% &\bf{90.5\%} &\bf{99.4\%} & 93.5\%\\
\hline
\end{tabular}
\end{center}
\label{Tab:com_map}
\end{table*}

To demonstrate the effectiveness of our translation-scale invariant image mapping and our data augmentation method, we compared with other skeleton data image mapping method and the results are presented in Table.\ref{tab:analysis}, which clearly demonstrates the effectiveness of our image mapping method. For a fair comparison, all the encoded action images are fine-tuned on the Alexnet \cite{krizhevsky2012imagenet}. and the result of Wang \etal is quoted from~\cite{Wang2016Action} directly, in order to avoid the hyper-parameter setting influence. It is worth noting that the results of Wang 
etal~\cite{Wang2016Action} utilized data augmentation. For fair comparison, we also give the results of our method with data augmentation.

It is clear from the Tab~\ref{Tab:com_map}, our mapping method outperform~\cite{Yong2015Skeleton} in all the dataset with a clearly margin, which prove our translation-scale property is important. In addtion, our method outperform~\cite{Wang2016Action} in most dataset except G3D. More importantly, our advantage is obvious.

\subsection{Effect of our multi-scale architecture}
We compared the performance of different pre-trained CNN net in Table \ref{tab:network}. It is obvious that our multi-scale network structure also improves the performance significately.

\begin{table}[htb]
\begin{center}
\caption{Performance of different pre-trained CNN model and our multi-scale network structure} \label{tab:network}
\begin{tabular}{  c  c  c }
\hline
Pre-trained CNN              &Cross subject &Cross view \\
\hline
AlexNet     & 81.3\% &87.2\% \\
VGGNet     & 84.6\% &90.1\% \\
ResNet101     & 83.8\% &89.9\% \\
ResNet152     & 84.3\% &90.1\% \\
ResNet101 + 2scale     & {84.6\%} &{90.9\%} \\
ResNet101 + 3scale     & {84.6\%} &{92.1\%} \\
ResNet152 + 3scale     & \bf{85.02\%} &\bf{92.3\%} \\
\hline
\end{tabular}
\end{center}
\end{table}

\subsection{Adapted to 2D skeleton}

As for 2D skeleton, we set the missing coordinate to 0. The correspondent results are presented in Tab.~\ref{tab:1D}. Surprisingly, the 2D skeleton is just a little worse than the 3D skeleton. We would like to argu that, to our best knowledge, we are the first to conduct this kind of experiment. This promissing results show that our method could also work well on the 2D skeleton data.

In addition, we also give the result of 1D skeleton data in Tab.~\ref{tab:1D}. The performance of 1D skeleton data decrease sharply.

\begin{table}[htb]
\begin{center}
\caption{Performance of 2D skeleton data. The experiments are conducted on NTU RGB+D dataset with fine-tuned on AlexNet} \label{tab:1D}
\begin{tabular}{  c  c  c }
\hline
coordinate              &Cross subject &Cross view \\
\hline
X-Y-Z     & 80.2\% &85.0\% \\
X-Y     & 77.8\% &83.5\% \\
Y-Z     & 77.4\% &83.9\% \\
X-Z     & 76.6\% &82.5\% \\
X     & {54.3\%} &{70.2\%} \\
Y     & {53.6\%} &{69.2\%} \\
Z     & {49.5\%} &{68.3\%} \\
\hline
\end{tabular}
\end{center}
\end{table}
%
%
\section{Conclusion and Future Work}
In this paper, we present a skeleton based action recognition method by using both translation-scale invariant image mapping and multi-scale deep CNNs. Experiments on the large scale challenging NTU RGB-D, UTD-MHAD MSRC-12 dataset show that our method outperforms the state-of-the-art methods by a large margin. In addition, we extend our method to 2D skeleton-based video recogntion problem and it performs well. 


\section*{Acknowledgements}
This work was supported in part by Natural Science Foundation of China grants (61420106007, 61671387) and Australian Research Council grants (DE140100180).
\bibliographystyle{../bibtex/IEEEtran}
\bibliography{CSRef}

\end{document}